\documentclass[a4paper, 10pt, conference]{ieeeconf}      
\usepackage{FG2024}
\usepackage[misc,geometry]{ifsym}
\usepackage{multirow}
\usepackage{makecell}
\usepackage{times}  
\usepackage{helvet}  
\usepackage{courier}  
\usepackage[hyphens]{url}  
\usepackage{graphicx} 
\usepackage{caption} 
\usepackage{pifont}
\newcommand{\cmark}{\ding{51}}%
\usepackage{algorithm}
\usepackage{algorithmic}
\usepackage{url}
\usepackage{amsmath}  
\usepackage{amsfonts,amssymb}
\FGfinalcopy 

\IEEEoverridecommandlockouts                              
\def\FGPaperID{173} 

\title{\LARGE \bf
Uncertainty-aware Bridge based Mobile-Former Network for Event-based Pattern Recognition
}

\author{\parbox{16cm}{\centering
    { Haoxiang Yang\dag$^1$, Chengguo Yuan\dag$^1$, Yabin Zhu$^{1,2}$, Lan Chen$^{2}$,  Xiao Wang$^{1}$, Futian Wang$^{1,3}$}\\
    {\normalsize
    $^1$ School of Computer Science and Technology, Anhui University, Hefei 230601, China  \\
    $^2$ School of Electronic and Information Engineering, Anhui University, Hefei 230601, China \\ 
    $^3$ Suzhou Glink IoT Technology Co., Ltd. } 
	\thanks{$\dag$~The first two authors contribute equally to this work. \\ 
	\Letter~Corresponding author: Lan Chen (chenlan@ahu.edu.cn), Xiao Wang (wangxiaocvpr@foxmail.com)}    
    }
  }

\begin{document}

\ifFGfinal
\thispagestyle{empty}
\pagestyle{empty}
\else
\author{Anonymous FG2024 submission\\ Paper ID \FGPaperID \\}
\pagestyle{plain}
\fi
\maketitle

\begin{abstract}
The mainstream human activity recognition (HAR) algorithms are developed based on RGB cameras, which are easily influenced by low-quality images (e.g., low illumination, motion blur). Meanwhile, the privacy protection issue caused by ultra-high definition (HD) RGB cameras aroused more and more people's attention. Inspired by the success of event cameras which perform better on high dynamic range, no motion blur, and low energy consumption, we propose to recognize human actions based on the event stream. We propose a lightweight uncertainty-aware information propagation based Mobile-Former network for efficient pattern recognition, which aggregates the MobileNet and Transformer network effectively. Specifically, we first embed the event images using a stem network into feature representations, then, feed them into uncertainty-aware Mobile-Former blocks for local and global feature learning and fusion. Finally, the features from MobileNet and Transformer branches are concatenated for pattern recognition. Extensive experiments on multiple event-based recognition datasets fully validated the effectiveness of our model. 
The source code of this work will be released at \textcolor{blue}{\url{https://github.com/Event-AHU/Uncertainty_aware_MobileFormer}}. 
\end{abstract}

\section{INTRODUCTION}
Human Activity Recognition (HAR) is one of computer vision's most critical tasks and developed significantly in recent years~\cite{ahmad2021graph, kong2018humanARSurvey} with the help of deep learning. Usually, these models are designed for video frames captured using RGB cameras and are widely used in practical applications. For example, we can achieve pre-prevention, in-process monitoring, and post-inspection in security monitoring field, intelligent referee in sports through the analysis of human behavior. Although the RGB cameras based HAR works well in simple scenarios, however, the issues caused by its imaging quality may limit the applications of HAR severely, such as low illumination and fast motion. On the other hand, the privacy protection is also widely discussed in the human-centered research. Awkwardly, \emph{the ethical problems} caused by high-quality data and \emph{the data quality problems} caused by low-quality video both require new behavior recognition paradigms.

Recently, the event camera (also termed Dynamic Vision Sensors, DVS) which is a bio-inspired sensor draws more and more attention from researchers~\cite{wang2023unleashing}~\cite{wang2023sstformer}~\cite{wang2023visevent}. Different from the RGB camera which records the scene into video frames in a synchronous way, each pixel in the event camera is triggered asynchronously by saving an event point if and only if the variation of intensity exceeds the given threshold. Due to the aforementioned unique imaging principle, the event camera shows the following advantages or features: \emph{high dynamic range}, \emph{low energy-consumption}, \emph{dense temporal resolution but sparse spatial resolution} \cite{gallegoevent}. Therefore, it performs well even in low-illumination, overexposure, and fast-motion scenarios. Also, the spatial resolution is getting higher, for example, $1280 \times 800$ and $1280 \times 720$ can be achieved by the CeleX-V~\cite{chen2019celexV} and PROPHESEE, respectively. These features all inspired us to address the pain points of HAR using an event camera.

In this work, we propose a new lightweight event-based recognition model by aggregating the MobileNet~\cite{howard2017mobilenets} and Transformer~\cite{khan2021transformers} networks based on an uncertainty-aware bridge module. The key insight of this work is that the CNN models the local features well and the Transformer performs better on the long-range relations mining. Unlike the typical dual-stream fusion network architecture, we feed the event data into the MobileNet branch and input random initialized tokens into the Transformer branch, as shown in Fig.~\ref{framework}. Inspired by Mobile-Former~\cite{chen2022mobileformer}, we propose an enhanced bridge module to connect dual parallel branches by considering the information propagation with uncertainty. The key insight of this idea is that the two branches focus on different types of feature learning, therefore, the information from different samples or
the same sample at different time steps may be asymmetrical. The decision of which branch should transmit richer information to the other one carries a certain level of uncertainty. Specifically, we model the uncertainty using the Gaussian distribution and adopt two MLP layers to predict its mean and variations. Then, the re-parameterization trick~\cite{doersch2016vae} is adopted to sample a new feature from the Gaussian distribution and fuse with another kind of feature. Extensive experiments validated the effectiveness of such uncertainty-aware information propagation modules between dual branches.

To sum up, the contributions of this paper can be summarized as the following three aspects: 

$\bullet$ We propose a novel lightweight uncertainty-aware Mobile-Former framework for event-based pattern recognition. It is a parallel dual-branch framework that simultaneously models the local and global features and effectively regulates the information flow between the dual branches.

$\bullet$ We propose a new uncertainty-aware bridge block which effectively boosts the feature interaction and fusion between the local CNN features and global Transformer features.

$\bullet$ Extensive experiments conducted on multiple widely used event-based recognition benchmark datasets fully demonstrate the effectiveness of the proposed model.

\section{RELATED WORK} 

In this section, we will introduce the related works about Event-based Recognition and Uncertainty-aware Learning. 

\noindent 
\subsection{Event-based Recognition} 
Current works can be divided into three streams for the event-based recognition, including the CNN based~\cite{wang2019evGait}, SNN based~\cite{fang2021PLIF, fang2021SNNIIR}, GNN based models~\cite{bi2019gnnevent, bi2020graph}, due to the flexible representation of event streams. 
For the CNN based models, Wang et al.~\cite{wang2019evGait} propose to identify human gaits using event camera and design a CNN model for recognition. 
As the third generation of neural networks, the SNN is also adopted to encode the event stream for energy-efficient recognition. To be specific, 
Peter et al.~\cite{diehl2015snnbalancing} propose the weight and threshold balancing method to achieve efficient ANN-to-SNN conversion. 
Nicolas et al.~\cite{perez2021sparse} propose a sparse backpropagation method for SNNs and achieve faster and more memory efficient.

For the point cloud based representation, Wang et al.~\cite{wang2019spaceCloud} treat the event stream as space-time event clouds and adopt PointNet~\cite{qi2017pointnet} as their backbone for gesture recognition. 
Sai et al. propose the event variational auto-encoder (eVAE)~\cite{vemprala2021representation} to achieve compact representation learning from the asynchronous event points directly. 
Fang et al.~\cite{fang2021snnresnet} propose SEW (spike-element-wise) residual learning for deep SNNs which addresses the vanishing/exploding gradient problems effectively. 
Meng et al.~\cite{meng2022DSR} propose an accurate and low latency SNN based on the Differentiation on Spike Representation (DSR) method. 
TORE~\cite{baldwin2022TORE} is short for Time-Ordered Recent Event (TORE) volumes, which compactly stores raw spike timing information. 
VMV-GCN~\cite{xie2022vmv} is proposed by Xie et al. which is a voxel-wise graph learning model to fuse multi-view volumetric. 
Li et al.~\cite{li2022eventFormer} introduce the Transformer network to learn event-based representation in a native vectorized tensor way. 
Different from these works, in this paper, we design a novel uncertainty-aware MobileFormer network that effectively aggregates the CNN and Transformer.

\noindent 
\subsection{Uncertainty-aware Learning}   
Uncertainty-aware learning is widely exploited in machine learning and computer vision tasks. Specifically, ~\cite{li2023uncertaintyPAR} propose a dual uncertainty-aware pseudo-labeling method for self-training to achieve knowledge transfer. \cite{wang2022uncerUDA} propose an uncertainty-aware clustering framework for unsupervised domain adaptive task. Qin et al.~\cite{qin2022uncerOSDA} adopts an uncertainty-aware method for federated Open set domain adaptation algorithm to generate a global model from all client models. Fang et al.~\cite{fang2023udnet} propose a novel uncertainty-aware salient object detection model, which use multiple supervision signals to teach the networks not only to focus on saliency regions but also pixels surrounding the contour of salient objects. Zhang et al.~\cite{zhang2021uncerQualityAssess} propose a unified blind image quality assessment (BIQA) model and also propose a hinge constraint to regularize uncertainty estimation when optimizing their model. Le et al.~\cite{le2023uncerExpressRecog} propose an uncertainty-aware label distribution learning approach for the improvement of the robustness of deep models against uncertainty and ambiguity for facial expression recognition. 
Different from existing works, we model the message propagation between CNN and Transformer networks using an uncertainty-aware learning approach which further improves the final recognition performance.

\section{METHODOLOGY}  

\subsection{Overview}  
As shown in Fig.~\ref{framework}, given the event streams, we first stack them into event images and extract their features using StemNet. The uncertainty-aware MobileFormer block is proposed and stacked as the backbone network. Specifically, we adopt the MobileNet as the CNN branch to extract the local features and utilize the Transformer to model the long-range relations. Note that, the Transformer takes the randomly initialized tokens as the input. To boost the interaction between the dual branches, we propose a novel uncertainty-aware bridge module to control the message passing. For the feature flows from CNN to the Transformer branch, we propose two MLPs to predict the Gaussian distribution using the output mean and variance values. Then, we sample a feature vector from the Gaussian distribution via re-parameterization tricks and aggregate it with the input of Transformer branch using cross-attention mechanism. Similar operations are conducted for the controllable information flow from the Transformer to CNN branch. The output of CNN and Transformer branches of the last uncertainty-aware Mobile-Former block are concatenated as the final feature representation. A classification head consisting of two dense layers is used for classification.

\subsection{Network Architecture}

\noindent 
\textbf{Input Representation and Embedding.~} 
Each point in the event stream $\mathcal{E}$ is usually represented as a tuple $e = \{x, y, t, p\}$, where $x, y$ are spatial coordinates, $t$ is the timestamp, and $p$ denotes the polarity (e.g., +1 and -1 denotes positive and negative event point). In this work, we stack the event streams into multiple event images due to their simplicity and effectiveness. Specifically, we first split the event streams into a fixed number of tensor tubes $T_i, i \in \{1, 2, ..., M\}$, then, each split is transformed into one event frame $F_i, i \in \{1, 2, ..., M\}$. The obtained event images are visualized in Fig.~\ref{fig:featureVIS}. 

After we get the event images, we resize them into a fixed resolution $224 \times 224$ and design a StemNet to project them into feature embeddings. Specifically, a 3D convolutional layer (kernel size: $3 \times 3 \times 3$) is proposed to achieve this embedding and get feature maps $F_{emb} \in \mathbb{R}^{4 \times 112 \times 112 \times 24}$.

\begin{figure}
\center
\includegraphics[width=3.3in]{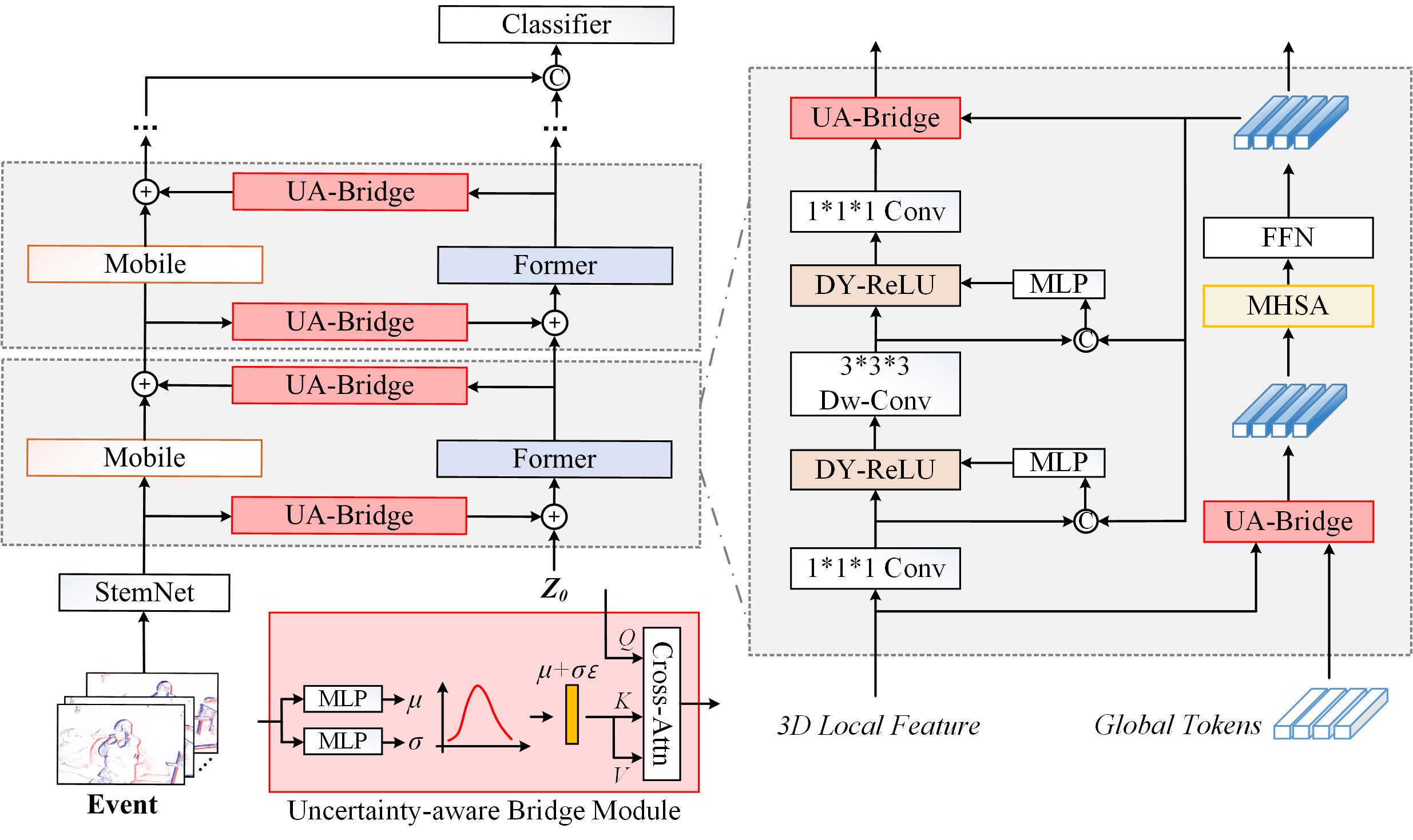}
\caption{An overview of our proposed uncertainty-aware bridge based Mobile-Former framework for event-based pattern recognition.}  
\label{framework}
\end{figure}

\noindent 
\textbf{Uncertainty-aware Bridge based Mobile-Former.~}
Given the feature embeddings of event streams, we design a novel uncertainty-aware Mobile-Former network that consists of lightweight mobile layers, Transformer layers, and uncertainty-aware bridge (UA-Bridge) modules. As shown in Fig.~\ref{framework}, the event feature embeddings are fed into a $1 \times 1 \times 1$ 3D convolutional layer and dynamic ReLU (DY-ReLU) layer. Then, two depth-wise 3D convolutional layers (kernel size: $3 \times 3 \times 3$) are utilized for local feature mining. The output will pass through a new DY-ReLU layer and two $1 \times 1 \times 1$ 3D convolutional layers. Note that, the local feature learning in the MobileNet branch also considers the information from the Transformer branch by dynamically updating the ReLU layer.

The input feature embeddings are also passed into the UA-Bridge module which contains two MLPs for Gaussian distribution estimation. This module will adaptively control the message propagation between the dual branches. Mathematically, the mean and the variance of the predicted message passing can be written as: 
\begin{equation}
    \label{meanMLP} 
    \mu = MLP_1(F_{emb}), ~~~~ \sigma = MLP_2(F_{emb}) 
\end{equation}
Then, a multi-variate Gaussian distribution can be built using the predicted mean and variance. The filtered features $F_{mf}$ from MobileNet to Transformer branch can be sampled from the Gaussian distribution using the reparameterization trick~\cite{kingma2015reparameterization}: 
\begin{equation}
    \label{sampling} 
    F_{mf} = \mu + \varepsilon * \sigma, ~~~~ \varepsilon \sim \mathcal{N}(0, I) 
\end{equation}
After that, we adopt a cross-attention layer to aggregate the $F_{mf}$ with Transformer tokens $Z_i, i = \{1, 2, ..., N\}$, which can be formulated as: 
\begin{equation}
    \label{crossAttention} 
    CrossAtten(Q, K, V) = softmax(\frac{QK^T}{\sqrt{d}})V. 
\end{equation}
where $d$ is the dimension of input feature vectors. For the cross-attention from MobileNet to Transformer, in our case, the $F_{mf}$ is reshaped into a feature vector and treated as Query ($Q$) and Key ($K$), and $Z_i$ is the Value ($V$). For the opposite direction, the global tokens of Transformer are used as the Query and Key and the local CNN features are the Value.

For the Transformer branch, we take the randomly initialized tokens as the input and fuse them with CNN features using a cross-attention layer as mentioned above. Then, the standard multi-head self-attention (MHSA) and feed-forward networks are adopted for long-range global feature learning. The output tokens will be used for parameter updating in the ReLU layer and also the UA-Bridge module for aggregation with CNN features. Similar operations are also conducted in subsequent Mobile-Former blocks.

\noindent 
\textbf{Classification Head and Loss Function.~} 
The local features and global features from CNN and Transformer blocks are concatenated and fed into two fully connected layers for pattern recognition. The cross entropy-loss function is adopted for the training of our whole framework in an end-to-end manner which can be formulated as: 
\begin{align}
Loss = -\frac{1}{B}\sum_{b=1}^{B}\sum_{n=1}^{N}Y_{bn}logP_{bn}
\end{align}
where $B$ denotes the batch size, $N$denotes the number of event classes. $Y$ and $P$ represent the ground truth and predicted class labels of the event sample, respectively.

\section{EXPERIMENTS} 

\subsection{Dataset and Evaluation Metric}  

In this paper, our experiments are conducted on the \textbf{ASL-DVS}~\cite{bi2020graph}, \textbf{N-Caltech101}~\cite{orchard2015converting}, \textbf{DVS128-Gait-Day}~\cite{wang2021eventgait3Dgraph} dataset. The \textbf{top-1} accuracy is adopted as the evaluation metric for the evaluation of our proposed model and other SOTA pattern recognition algorithms.

\subsection{Implementation Details} 
Our proposed lightweight uncertainty-aware bridge based Mobile-Former framework can be optimized in an end-to-end manner. The learning rate and weight decay are set as 0.0001 and 0.1, respectively. The AdamW~\cite{loshchilov2023adamw} is selected as the optimizer and trained for a total of 60 epochs. In our implementations, 12 blocks are stacked as our backbone network. For the input of the Transformer branch, we randomly initialized 6 tokens and also tested other settings which will be discussed in experiments. We select 8 event frames as the input of MobileNet branch. Our code is implemented based on PyTorch~\cite{NEURIPS2019_9015} framework and the experiments are conducted on a server with GPU V100.

\subsection{Comparison with Other SOTA Algorithms}

\noindent 
\textbf{Results on ASL-DVS~\cite{bi2020graph}.~}
As shown in Table~\ref{ASLDVSResults}, our proposed method achieves $99.9\%$ on this benchmark dataset which is a new SOTA performance. The compared method M-LSTM which adopts learnable event representation is still inferior to our method. Some graph-based event recognition models are also worse than ours, including EV-VGCNN, VMV-GCN, and Ev-Gait-3DGraph. Therefore, we can draw the conclusion that our proposed lightweight model is effective for event-based pattern recognition.

\begin{table}[!htp]
\center

\scriptsize   
\caption{Results on the ASL-DVS dataset.} 
\label{ASLDVSResults}
\begin{tabular}{ccccc} 		
\hline 
\makecell[c]{\textbf{EST} \\ \cite{gehrig2019end}} &\makecell[c]{\textbf{AMAE} \\ \cite{deng2020amae}}     &\makecell[c]{\textbf{M-LSTM}\\ \cite{behera2018context}}    &\makecell[c]{\textbf{MVF-Net}\\ \cite{deng2021mvf}}      & \makecell[c]{\textbf{EventNet}\\ \cite{sekikawa2019eventnet}} \\  
\hline 
0.979   & 0.984     &0.980     &0.971     &0.833    \\ 
\hline 
\makecell[c]{\textbf{RG-CNNs} \\ \cite{bi2020graph}}   &\makecell[c]{\textbf{EV-VGCNN}\\ \cite{deng2021evvgcnn}}   &\makecell[c]{\textbf{VMV-GCN}\\ \cite{xie2022vmvgcn}}   &\makecell[c]{\textbf{EV-Gait-3DGraph}\\ \cite{wang2019ev}}    &\textbf{Ours} \\
\hline 
0.901     &0.983     &0.989  &0.738  &0.999	 \\
\hline
\end{tabular}
\end{table}

\noindent 
\textbf{Results on N-Caltech101~\cite{orchard2015converting}.~} 
As shown in Table~\ref{Caltech101Results}, our model achieves 0.798 on the top-1 accuracy on this benchmark dataset which is significantly better than the compared methods. To be specific, our model outperforms the ResNet50 by $+16.1\%$ on the top-1 metric and beats the VMV-GCN by $+2\%$ which ranks the second place. Thanks to the uncertainty-aware local and global feature learning, our model achieves superior performance which fully validated its effectiveness.

\begin{table}[!htp]
\center
\scriptsize  
\caption{Results on N-Caltech101 dataset.} 
\label{Caltech101Results} 
\setlength\tabcolsep{3.3pt}
\begin{tabular}{ccccccc} 		
\hline
\makecell[c]{\textbf{EventNet}\\ \cite{sekikawa2019eventnet}}   &\makecell[c]{\textbf{Gabor-SNN}\\ \cite{sironi2018hats}}     &\makecell[c]{\textbf{RG-CNNs}\\ \cite{bi2020graph}}    &\makecell[c]{\textbf{VMV-GCN}\\ \cite{xie2022vmvgcn}}     &\makecell[c]{\textbf{EV-VGCNN}\\ \cite{deng2021evvgcnn}}     &\makecell[c]{\textbf{EST}\\ \cite{gehrig2019end}}     	  		 \\
0.425    &0.196     &0.657     &0.778   &0.748     &0.753     	  		 \\
\hline
\makecell[c]{\textbf{ResNet-50}\\ \cite{mascarenhas2021comparison}}  &\makecell[c]{\textbf{MVF-Net}\\ \cite{deng2021mvf}}     &\makecell[c]{\textbf{M-LSTM}\\ \cite{behera2018context}}     &\makecell[c]{\textbf{AMAE}\\ \cite{deng2020amae}}     &\makecell[c]{\textbf{HATS}\\ \cite{sironi2018hats}}     &\textbf{Ours}     	  		 \\ 
0.637    &0.687     &0.738     &0.694     &0.642 &0.798     	  		 \\
\hline
\end{tabular}
\end{table}

\noindent 
\textbf{Results on DVS128-Gait-Day~\cite{wang2021eventgait3Dgraph}.~} 
This dataset is specifically proposed for human gait recognition by Wang et al., as shown in Table~\ref{DVS128GaitResults}, we can find that the EVGait-3DGraph already achieves $94.9\%$ on this dataset. In contrast, our proposed method obtains $95.9\%$ which is better than this 3D graph based recognition model. The superior results on this dataset fully demonstrate that our model works well on event-based recognition.

\begin{table}[!htp]
\center
\scriptsize   
\caption{Results on the DVS128-Gait-Day dataset.} 
\label{DVS128GaitResults} 
\begin{tabular}{ccccccc} 		
\hline 
\textbf{EVGait-3DGraph}~\cite{wang2019ev}   &\textbf{2DGraph-3DCNN}~\cite{bi2020graph} &\textbf{EV-Gait-IMG}~\cite{wang2021event} \\ 
94.9   &92.2     &87.3 \\
\textbf{LSTM-CNN}~\cite{xia2020lstm}   &\textbf{SVM-PCA}~\cite{khamparia2021svm}   &\textbf{Ours}  \\
86.5     &78.05     &95.9     	  		 \\
\hline 
\end{tabular}
\end{table}

\subsection{Ablation Study} 

To help the readers better understand the effectiveness of aggregating MobileNet and Transformer branches, as shown in Table~\ref{CAResults}, we isolate these components separately to validate their effectiveness. To be specific, when only MobileNet or Transformer is used for recognition, the results are $76.53\%$ and $58.01\%$, respectively. When both branches are aggregated using the cross-attention mechanism, the results can be improved to $76.83\%$, which demonstrates that it is effective when fusing the local and global features for event-based recognition. 
For the ReLU activation layer, it is possible to predict its parameters using MLPs in an online manner to attain greater flexibility and robustness. We can find that the results can be improved to $77.94\%$ when the dynamic ReLU (DY-ReLU) is adopted.

In this work, we exploit the uncertainty-aware message propagation between MobileNet and Transformer branch to achieve effective feature aggregation. As shown in Table~\ref{CAResults}, we can find that the results can be improved to $79.80\%$ when this module is adopted. This experiment fully validated the effectiveness of controllable information propagation for event-based recognition.

\begin{table}[!htp]
\scriptsize 
\tiny       
\caption{Component analysis on N-Caltech101 dataset. UAB and CA denote the Uncertainty-aware Bridge and Cross Attention module, respectively.}  
\label{CAResults} 
\resizebox{\columnwidth}{!}{
\begin{tabular}{c|ccccc|c} 		
\hline 
\textbf{No.}  & \textbf{Mobile} &\textbf{Former} &\textbf{UAB}  &\textbf{CA}  &\textbf{DY-ReLU} &\textbf{Results}   \\   
\hline 
1 &\cmark   &        &               &   &   &76.53   \\
2 &         &\cmark   &              &        & &58.01      \\
3 &\cmark   &\cmark   &              &\cmark        & &76.83       \\
4 &\cmark   &\cmark  &              &\cmark        &\cmark &77.94       \\
5 &\cmark   &\cmark  &\cmark         &\cmark        &\cmark  &79.80       \\
\hline 
\end{tabular}} 
\end{table}

\subsection{Parameter Analysis} 
\begin{figure}
\center
\includegraphics[width=3.5in]{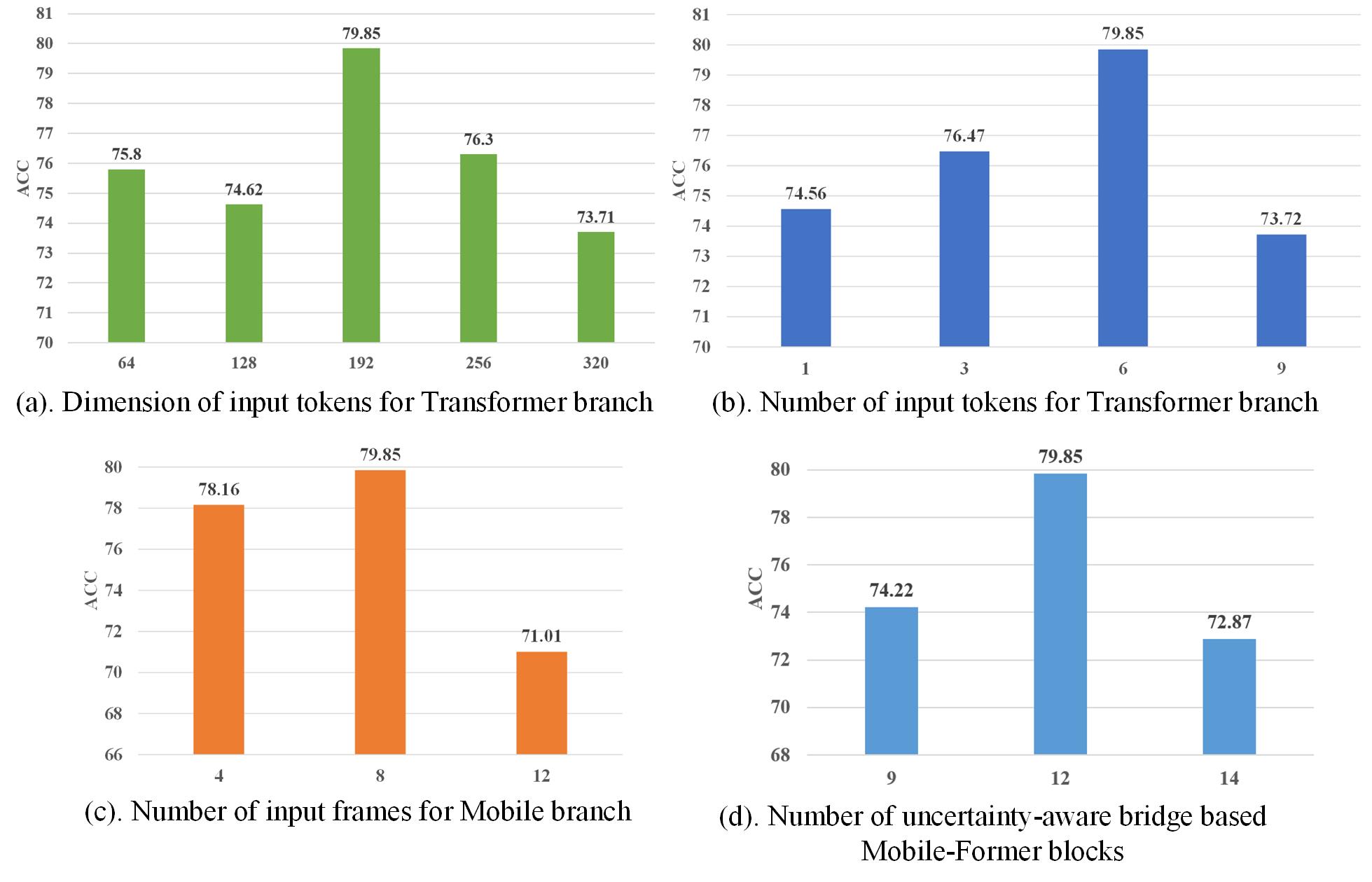}
\caption{Ablation study on parameter analysis.}  
\label{paramAnalysis}
\end{figure}

When building our network, there are multiple flexible design choices that can enhance the final recognition results. This section will explore the following aspects and report the corresponding results on N-Caltech101 dataset in Fig.~\ref{paramAnalysis}.

\noindent 
\textit{1). Number and Dimension of input tokens for Transformer:} We changed the dimension of input tokens into multiple versions, including 64, 128, 192, and 256. From Fig.~\ref{paramAnalysis} (a),  it is easy to find that the best performance can be achieved when the 192 is used. For the number of input tokens, we set it as 1, 3, 6, 9, and the corresponding results are 74.56, 76.47, 79.85, 73.72 on the top-1 accuracy, as shown in Fig.~\ref{paramAnalysis} (b).

\noindent 
\textit{2). Number of input frames for MobileNet:} We try to input 4, 8, and 12 event frames into the MobileNet branch, and the corresponding results are 78.16, 79.85, and 71.01, as shown in Fig.~\ref{paramAnalysis} (c). One can find that a better result can be obtained when eight frames are used for MobileNet. Introducing more video frames can bring in a substantial amount of redundant information, making it challenging for the model to extract useful features. An excessive number of frames might lead to model complexity, making training more difficult and potentially increasing the risk of overfitting. This phenomenon is also referred to as information overload.

\noindent 
\textit{3). Number of uncertainty-aware bridge based Mobile-Former blocks:} When building our backbone network, different uncertainty-aware Mobile-Former layers can be used. In this part, we exploit 9, 12, and 14 layers and get the top-1 results 74.22, 79.85, 72.87, as shown in Fig.~\ref{paramAnalysis} (d).

\subsection{Visualization}  

\begin{figure}
    \centering
    \includegraphics[width=\linewidth]{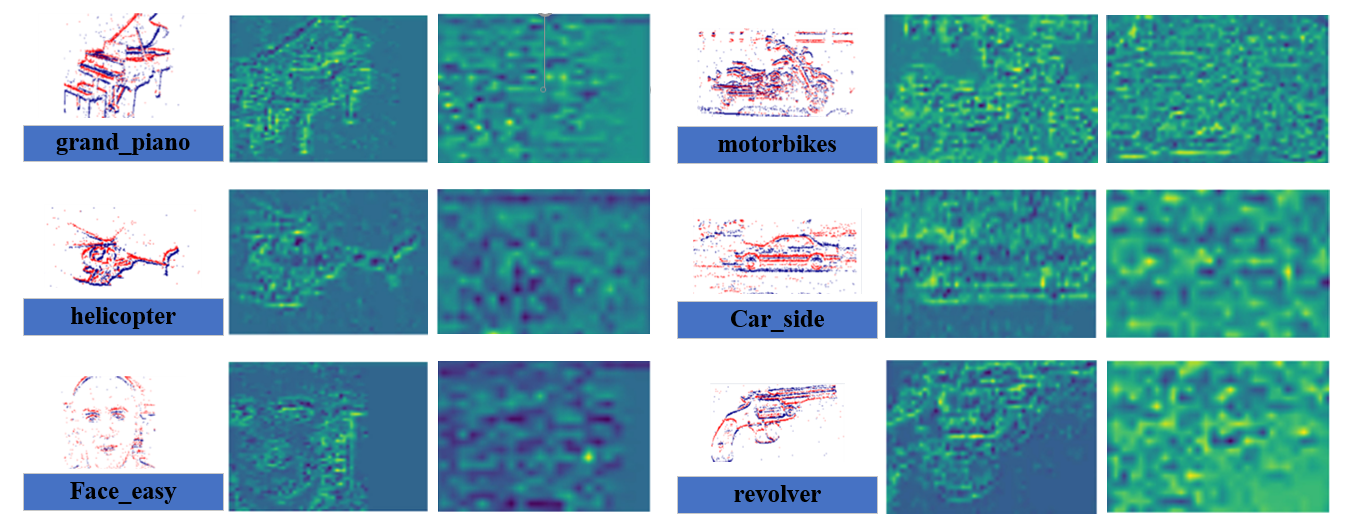}
    \caption{Visualization of the CNN features using our model on the N-Caltech101 dataset.}
    \label{fig:featureVIS}
\end{figure}

\begin{figure}
    \centering
    \includegraphics[width=\linewidth]{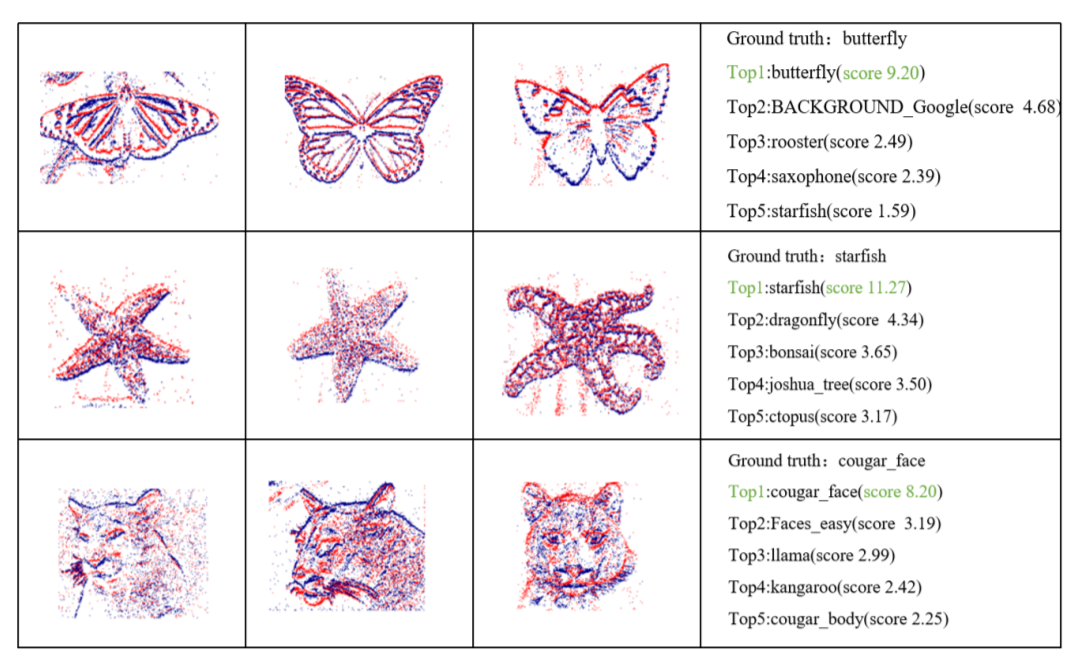}
    \caption{Visualization of the top-5 predicted categories using our model on the N-Caltech101 dataset.}
    \label{fig:top5results}
\end{figure}

\begin{figure}[!htp]
    \centering
    \includegraphics[width=\linewidth]{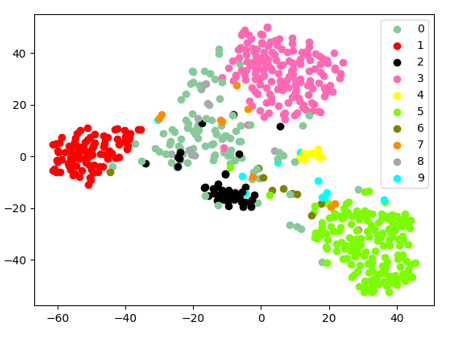}
    \caption{Visualization of our feature distribution on N-Caltech101.}
    \label{fig:distributionVIS}
\end{figure}

In addition to the aforementioned quantitative experimental analysis, we also provide visualizations to better assist readers in comprehending the effectiveness of our model. As shown in Fig.~\ref{fig:featureVIS}, we first give a visualization of MobileNet feature maps. One can find that our model performs well in capturing the active event regions. 
As shown in Fig.~\ref{fig:top5results}, we also present the top-5 recognition results and their corresponding confidence scores for model predictions. It is evident that our approach can accurately identify the patterns captured by the event camera. 
As shown in Fig.~\ref{fig:distributionVIS}, we randomly select 10 classes of samples from N-Caltech101 dataset and project the features into 2D spaces. It is easy to find that our model performs well in separating these categories.

\section{Conclusion}
In this work, we propose to recognize objects based on the event streams. we propose a lightweight uncertainty-aware information propagation based Mobile-Former network for efficient pattern recognition, which aggregates the MobileNet and Transformer network effectively. Extensive experiments on multiple event-based recognition datasets fully validated the effectiveness of our model. In our future works, we will consider a knowledge distillation strategy to further enhance the final recognition performance.

{\small
\bibliographystyle{ieee}

\bibliography{egbib}
}

\end{document}